\documentclass[sigconf, final]{acmart}
\AtBeginDocument{%
  \providecommand\BibTeX{{%
    \normalfont B\kern-0.5em{\scshape i\kern-0.25em b}\kern-0.8em\TeX}}}

\setcopyright{acmcopyright}
\copyrightyear{2025}
\acmYear{2025}
\acmDOI{10.1145/XXXXXXX}

\acmConference[MORE '25]{Sydney '25: ACM Web Conference}{28 April-2 May 2025}{Sydney, Australia}
\acmBooktitle{Woodstock '18: ACM Symposium on Neural Gaze Detection, June 03--05, 2018, Woodstock, NY}
\acmPrice{15.00}
\acmISBN{978-1-4503-XXXX-X/18/06}

\acmSubmissionID{3}

\usepackage{times}
\usepackage{epsfig}
\usepackage{graphicx}
\usepackage{amsmath}

\usepackage{amssymb}

\usepackage{algorithm, algpseudocode}
\usepackage{multirow}
\usepackage{caption}
\usepackage{comment}
\usepackage{amsmath}
\usepackage{amssymb}
\usepackage{CJKutf8}
\usepackage{booktabs}
\definecolor{citecolor}{RGB}{119,185,0} 

\usepackage{pifont}
\usepackage{color}

\newlength\savewidth

\begin{document}
\begin{CJK}{UTF8}{gbsn}
\title{Efficient Vision Language Model Fine-tuning for Text-based Person Anomaly Search}

\author{Jiayi He \quad Shengeng Tang \quad Ao Liu \quad Lechao Cheng \quad Jingjing Wu\quad Yanyan Wei}
\affiliation{%
   School of Computer Science and Information Engineering, Hefei University of Technology\\
}
\email{hejy4396@mail.hfut.edu.cn, {tangsg, chenglc, wujingjing, weiyy}@hfut.edu.cn, xzliuao@yau.edu.cn}
\thanks{Shengeng Tang is the corresponding author.}

\begin{abstract}
This paper presents the HFUT-LMC team's solution to the WWW 2025 challenge on Text-based Person Anomaly Search (TPAS). The primary objective of this challenge is to accurately identify pedestrians exhibiting either normal or abnormal behavior within a large library of pedestrian images. Unlike traditional video analysis tasks, TPAS significantly emphasizes understanding and interpreting the subtle relationships between text descriptions and visual data. The complexity of this task lies in the model's need to not only match individuals to text descriptions in massive image datasets but also accurately differentiate between search results when faced with similar descriptions. To overcome these challenges, we introduce the Similarity Coverage Analysis (SCA) strategy to address the recognition difficulty caused by similar text descriptions. This strategy effectively enhances the model's capacity to manage subtle differences, thus improving both the accuracy and reliability of the search. Our proposed solution demonstrated excellent performance in this challenge.
\end{abstract}



\maketitle

\section{Introduction}
Text-based Person Anomaly Search (TPAS)~\cite{yang2024beyond} is an extension and development of Text-based Person Search (TPS)~\cite{li2017person, lin2023diff, park2025plot, cao2024empirical, bai2023text}, which is closely related to some classic cutting-edge tasks such as cross-modal retrieval~\cite{wang2019position, jiang2024cala, wang2020pfan++}, person re-identification~\cite{wu2024comprehensive, wu2024intermediary, wang2025MORE, 10.1145/3581783.3612009}, visual-linguistic analysis~\cite{tang2022graph, song2024emotional, tang2022gloss, NEURIPS2021}, and cross-media reasoning~\cite{tang2024sign, ma2024modality, tang2024gloss}. The primary task of TPS involves using text descriptions as queries to retrieve matching pedestrian images from a large image library. Traditional TPS works mainly focus on identifying the appearance features of pedestrians. However, TPS frequently overlooks the importance of motion information. TPAS not only emphasizes pedestrian appearance features but also targets the identification of behavioral abnormalities. It aims to accurately identify pedestrians exhibiting normal or abnormal behavior from numerous candidates based on specified appearance and action descriptions.

Compared to TPS, TPAS encounters more complex challenges. A primary challenge is bridging the semantic divide between text and images. This divide arises from the distinct modalities through which text and images convey information. In these cases, the model must meticulously parse and comprehend subtle yet critical differences in descriptions to ensure high-precision recognition. This challenge requires the model to achieve precise cross-modal semantic alignment between text and images and to identify subtle differences in appearance and actions. Another challenge occurs when processing text descriptions with similar content, leading the model to struggle to distinguish results clearly. Even with minor wording differences, corresponding images may appear similar, complicating the model's differentiation ability. This necessitates a deep understanding of the complex relationships between text and images, ensuring accuracy and consistency in search results when handling similar descriptions.

To enhance the model's accuracy and reliability in handling similar descriptions, we introduced the Similarity Coverage Analysis (SCA) strategy. The SCA strategy addresses this issue by comparing confidence levels of identical answers within similar text descriptions. The main contributions of our method are as follows:
\begin{itemize}
\item We fine-tuned X-VLM~\cite{zeng2022multigrainedvisionlanguagepretraining} with the PAB~\cite{yang2024beyond} dataset, thereby enhancing its discriminatory and recognition abilities relative to the baseline model.

\item We introduced the Similarity Coverage Analysis strategy to address the challenge of models struggling to differentiate between similar answers.

\item In the Text-based Person Anomaly Search challenge, our solution achieved a Recall@1 score of 85.49 on the test set. The experimental results demonstrate the effectiveness of our model.
\end{itemize}
 
\section{Related Work} \label{sec:realted work}
Text-based Person Anomaly Search better addresses real-world requirements by considering both appearance and action descriptions. This approach allows for the precise identification of target pedestrians exhibiting normal or abnormal behaviors among numerous candidates. Here, we review related techniques: Text-based Person Search, Person Anomaly Detection, and Vision Language Pre-training.

\subsection{Text-based Person Search}
Text-based Person Search~\cite{li2017person,zhang2018deep,niu2020improving} is a research field combining pedestrian re-identification and cross-modal retrieval, with the aim of retrieving pedestrian images using natural language descriptions. The core of TPS involves feature extraction and semantic alignment: discriminative features are extracted through pre-processing and end-to-end frameworks, while semantic alignment is achieved via cross-modal attention mechanisms and generative methods. Although TPS technology has significantly advanced in recent years, it still encounters challenges like modal heterogeneity in practical applications.

\subsection{Person Anomaly Detection}
Anomaly detection is a crucial issue garnering significant attention across numerous research and application domains, particularly in safety. Person anomaly detection, specifically tailored for safety, focuses on identifying and analyzing activities that deviate from normal behavior patterns to enhance safety and responsiveness. Currently, most pedestrian anomaly detection methods predominantly rely on video data rather than images, aiming to identify abnormal behavior events. This research is addressed as a one-class classification problem \cite{feng2021convolutional, flaborea2023multimodal, hirschorn2023normalizing, zaheer2022generative}, an unsupervised learning challenge \cite{zaheer2022generative}, or a supervised/weakly supervised issue \cite{zaheer2022generative, acsintoae2022ubnormal}.

\begin{figure*}[tbh]
  \centering
  \includegraphics[width=0.85\textwidth]{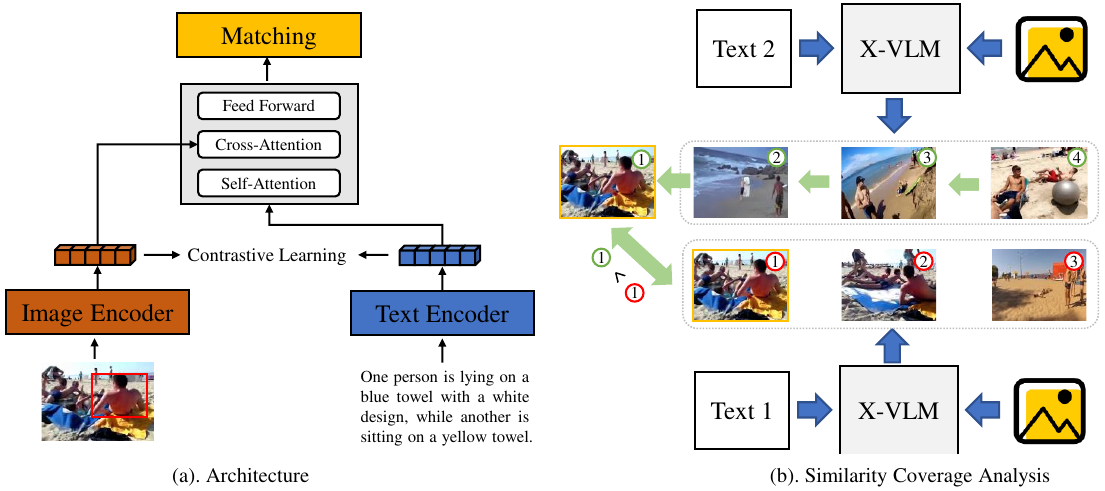}
  \caption{(a)X-VLM for Text-based Person Anomaly Search. (b)When two similar text descriptions with different answers yield the same result, compare the confidence scores for these answers. Replace the answer with the lower confidence score by using the answer from the group where the confidence score is lower than the current score.}
  \label{fig: Overall}
\end{figure*}

\subsection{Vision Language Pre-training}
Visual Linguistic Pre-training (VLP)~\cite{chen2023vlp, wei2025leveraging} is a collaborative training model integrating vision and language. It aims to extract rich visual and semantic features from large-scale multimodal data by simultaneously learning visual and linguistic tasks. To enhance image-text association learning and improve multimodal comprehension, single-stream~\cite{li2020oscar, zhou2020unified, zhang2020devlbert} and dual-stream architectures~\cite{dou2022empirical} are designed to process and fuse visual and verbal information differently. Additionally, to better capture semantic relationships between vision and language, researchers have proposed various pre-training objectives, such as masked language modeling~\cite{NEURIPS2021_50525975}, visual-language matching~\cite{li2020hero}, and other task-specific objectives~\cite{zhou2020unified, jiang2020defense}. Notably, VLP advancement relies on large-scale, high-quality datasets. Consequently, with continuous technological advancements and improved dataset quality, VLP is poised to excel in multimodal understanding, generative tasks, and practical application performance.

\section{Method} \label{sec:method}
\subsection{Task Definition}
Text-based Person Anomaly Search is a very challenging search task. Given a text description $T$, identify a subset $P_{T}$ of anomalous behavior individuals from the pedestrian image collection $P$ that matches the description. The formula is as follows:
\begin{eqnarray}
    P_{T} = \{P_i \in P | \arg\max[sim(F_{T},F_{P_{i}})]\},
\end{eqnarray}
where $sim$ represents the cosine similarity between text feature $F_{T}$ and image feature $F_{P_{i}}$, which is used to measure the degree of matching between the two.

\subsection{Overall Architecture}
Figure~\ref{fig: Overall} (a) illustrates the overall architecture of our method. Our method comprises three key components: an image encoder, a text encoder, and a cross-modal encoder. Initially, an image encoder extracts image features $F_{P}$ from input images, while a text encoder extracts text features $F_{T}$ from text data. These encoders focus on extracting information from their respective modalities to ensure the acquisition of high-quality image and text features. The formula is as follows:
\begin{eqnarray}
    F_{T} = TextEncoder(T), F_{P} = ImageEncoder(P).
\end{eqnarray}

Next, extracted image features $F_{P}$ are used as conditional information for the cross-modal encoder. The cross-modal encoder fuses and aligns information from different modalities to accurately understand the correlation and matching between them. Finally, the method predicts matching results between image and text using the cross-modal encoder's output. 
The formula is as follows:
\begin{eqnarray}
    Y^{match} = CMEncoder(F_{T},F_{P}),
\end{eqnarray}
where $CMEncoder$ denotes the cross-modal encoder.

\noindent\textbf{Contrastive Loss}
We predict image-text pairs, denoted as $(P_i,T)$, using in-batch negatives. A mini-batch of $N$ pairs is randomly sampled, and both image-to-text and text-to-image similarities are computed within the batch. For each pair $(P_i,T)$, where $T$ is the positive example for $I$, the remaining $N - 1$ texts within the mini-batch are treated as negative examples. The cosine similarity $sim(F_T,F_{P_i})$ is defined as:
\begin{eqnarray}
    sim(F_T,F_{P_i}) = F_{T}^{\top}F_{P_i}
\end{eqnarray}

The in-batch image-to-text similarity is then calculated as:
\begin{eqnarray}
    sim_{P \to T} = \frac{\exp(sim(P,T))}{\sum_{i=1}^{N} \exp(sim(P, T_i))}
\end{eqnarray}

where $T_i$ represents the text embeddings in the mini-batch, including both positive and negative examples. Similarly, the text-to-image similarity is calculated as:
\begin{eqnarray}
    sim_{T \to P} = \frac{\exp(sim(T,P))}{\sum_{i=1}^{N} \exp(sim(T, P_i))}
\end{eqnarray}

Finally, we define the contrastive loss for a single pair $(P,T)$ as the sum of the image-to-text and text-to-image similarities:
\begin{eqnarray}
    \mathcal{L}_{\text{contrastive}} = \frac{1}{2} \mathbb{E}_{P, T \sim D} \Big[
    \mathcal{H}(\mathbf{y}_{P \to T}(P), sim_{P \to T}(P)) \nonumber \\
    + \mathcal{H}(\mathbf{y}_{T \to P}(T), sim_{T \to P}(T))\Big]
\end{eqnarray}

The final loss function combines the contrastive loss with any additional regularization or fine-tuning losses, ensuring that both vision and text features are aligned in the shared latent space.  During training, this loss is minimized to learn better alignments between the image and text modalities.

\noindent\textbf{Match Loss}
We determine whether a pair of image-text is matched. For each image in a mini-batch, we sample an in-batch hard negative text based on $sim_{P \to T}$. Texts more relevant to the concept are more likely to be selected as negatives. Additionally, for each text, we sample one hard negative image. The matching probability, $sim_{\text{match}}$, is predicted using $x_{\text{cls}}$, the output [CLS] embedding from the cross-modal encoder. The loss is defined as:
\begin{eqnarray}
    \mathcal{L}_{\text{match}} = \mathbb{E}_{P, T \sim D}\mathcal{H}(y_{\text{match}}, 
    sim_{\text{match}}(P, T))
\end{eqnarray}
where $y_{\text{match}}$ is a 2-dimensional one-hot vector representing the ground-truth label.

\subsection{Similarity Coverage Analysis}
Discriminating between two images corresponding to similar text descriptions is a challenging problem. When two or more similar text descriptions $m$ correspond to different answers but yield identical search results, the accuracy of the final outcome may be compromised. To tackle this challenge, we propose the Similarity Coverage Analysis (SCA). As shown in Figure~\ref{fig: Overall} (b), similar text descriptions, Text $1$ and Text $2$, retrieve the same initial result, but only one is correct. When the model retrieves $n$ results for each text description, each result is associated with a probability value $s_i$. The formula is as follows:
\begin{eqnarray}
    R_i = \{(r_{(i,j)},s_{(i,j)})|i \in m, j \in n \}.
\end{eqnarray}
To enhance result accuracy, we employ the following strategy: Compare the same answers in the two sets of results and calculate $\Delta s$. The formula is as follows:
\begin{eqnarray}
    \Delta s = s_{(1,j)}-s_{(2,j)}.
\end{eqnarray}
Replace the result of $j$ with the result of $j+1$ according to the result of $\Delta s$. The SCA enables us to prioritize the most representative and accurate answers among multiple groups of similar results, thereby improving the model’s overall performance and reliability.

\section{Experiments} \label{sec:experiments}
\subsection{Dataset and Evaluation Metric}
\noindent\textbf{Dataset} Pedestrian Anomaly Behavior (PAB)~\cite{yang2024beyond} is an extensive image-text dataset that emphasizes pedestrians' abnormal behaviors. The dataset encompasses a variety of behaviors including running, performing, and playing football, alongside opposing abnormal behaviors like lying down, being hit, and falling. The training set comprises 1,013,605 pairs of synthetic normal and abnormal image-text pairs, whereas the test set includes 1,978 pairs of real-world image-text pairs. 

\noindent\textbf{Evaluation Metric} $Recall@k$ is a widely used evaluation metric in information retrieval, measuring the proportion of relevant items or positive examples successfully retrieved in the top $k$ recommendations. This metric is crucial for assessing a system's capability to identify as many relevant items or positive examples as possible within a limited set of recommendations. The formula is expressed as follows:
\begin{eqnarray}
    Recall@k=\frac{\lvert R \cap S_{k} \rvert}{ \lvert R \rvert}
\end{eqnarray}
where $R$ denotes the set of all relevant items, $S_{k}$ represents the set of the top $k$ items returned by the system, $ \lvert R \cap S_{k} \rvert$ is the number of relevant items found in the top $k$ results, and $ \lvert R \rvert$ is the total number of relevant items.

\subsection{Implementation Details}
We use the X-VLM model to fine-tuning the task of Text-based Person Anomaly Search. Specifically, image features are extracted using the pre-trained X-VLM encoder, which processes both visual and textual modalities. Input images are resized to 384×384 pixels during preprocessing before input into the model. A text-query embedding is generated by the pre-trained text encoder and is subsequently matched with the image feature space during retrieval. During fine-tuning, the maximum input length for text queries is limited to 70 tokens to ensure consistent text representation across the dataset. The batch size is configured to 16. The Adam optimizer is used for optimization with a learning rate of 3e-5, a weight decay of 0.01, and a learning rate multiplier of 2. The learning rate follows a linear decay schedule, starting from 3e-5 without any warm-up steps. The model is trained over 10 epochs.

\subsection{Experimental Results}
As shown in Table~\ref{tab: result}, we summarize the results of all participating teams on the PAB test set. Our team ultimately secured fourth place. Although a gap remains compared to the top teams, it is worth noting that our $Recall@1$ reached $85.49$, significantly outperforming the other two participating teams. This outcome reflects the effectiveness and innovation of our approach and indicates directions for future improvement and enhancement.

\begin{table}[tbp]
\renewcommand\arraystretch{1.1}
\caption{The results of Text-based Person Anomaly Search on the PAB test set.}
\vspace{-2mm}
\label{tab: result}
\centering
\resizebox{0.35\textwidth}{!}{
\begin{tabular}{cccc}
\bottomrule[1pt]
   \multicolumn{1}{c}{Team} &R@1{$\uparrow$} &R@5{$\uparrow$} &R@10{$\uparrow$} \\
\bottomrule[0.5pt]
   \multicolumn{1}{c}{DG}  &89.28 &99.34  &99.75 \\
   \multicolumn{1}{c}{AIO\_GenAI4E}  &89.23 &99.70 &99.85 \\
   \multicolumn{1}{c}{WeTryOurBest}  &89.18  &99.70  &99.85 \\
\multicolumn{1}{c}{\textbf{HFUT-LMC}}  &\bf{85.49}  &\bf{99.29}  &\bf{99.80}\\
\multicolumn{1}{c}{ctyun-ai}  &69.87 &95.40  &97.42\\
\multicolumn{1}{c}{CUFE}  &69.87 &95.40  &97.42\\
\bottomrule[1pt]
\end{tabular}}
\end{table}

\begin{table}[tbp]
\renewcommand\arraystretch{1.1}
\caption{The ablation results of the model. \textbf{Zero-shot}: Untrained, \textbf{Fine-tuning}: Trained.}
\vspace{-2mm}
\label{tab:ablation modules}
\centering
\resizebox{\linewidth}{!}{
\begin{tabular}{ccccccccc}
\bottomrule[1pt]
   \multicolumn{1}{l}{\multirow{2}{*}{Methods}} & \multicolumn{3}{c}{Zero-shot} & ~ & \multicolumn{3}{c}{Fine-tuning} \\
   \cline{2-4}\cline{6-8}
   \multicolumn{1}{c}{} &R@1{$\uparrow$} &R@5{$\uparrow$} &R@10{$\uparrow$} & ~ &R@1{$\uparrow$} &R@5 {$\uparrow$} &R@10{$\uparrow$}\\
\bottomrule[0.5pt]
   \multicolumn{1}{l}{EVA-CLIP~\cite{sun2023eva}}  &60.01 &90.14  &94.54  & ~ &77.81 &98.03  &99.39 \\
   \multicolumn{1}{l}{LLM2CLIP~\cite{huang2024llm2clippowerfullanguagemodel}}  &73.61 &96.97 &98.58 & ~ &79.93 &\bf{99.09} &\bf{99.75}\\
   \bottomrule[0.5pt]
   \multicolumn{1}{l}{X-VLM~\cite{zeng2022multigrainedvisionlanguagepretraining}}  &\bf{77.86}  &\bf{98.28}  &\bf{99.24}  & ~ &\bf{79.98}  &98.94  &\bf99.44 \\
\bottomrule[1pt]
\end{tabular}}

\end{table}

\begin{table}[tbp]
\renewcommand\arraystretch{1.1}
\caption{The ablation result with training data volume.}
\vspace{-2mm}
\label{tab: data volume}
\centering
\resizebox{0.28\textwidth}{!}{
\begin{tabular}{ccccc}
\bottomrule[1pt]
   \multicolumn{1}{c}{Methods} &R@1{$\uparrow$} &R@5{$\uparrow$} &R@10{$\uparrow$} \\
\bottomrule[0.5pt]
   \multicolumn{1}{c}{1W}    &\bf{85.49}  &\bf{99.29}  &\bf{99.80} \\
   \multicolumn{1}{c}{2W}    &84.98 &99.29 &99.80 \\
   \multicolumn{1}{c}{3W}    &84.43 &98.89  &99.44 \\
   \multicolumn{1}{c}{4W}    &83.52 &98.58  &99.54 \\
   \multicolumn{1}{c}{ALL}  &83.52 &98.84  &99.54 \\
\bottomrule[1pt]
\end{tabular}}
\vspace{-2mm}
\end{table}

\begin{table}[tbp]
\renewcommand\arraystretch{1.1}
\caption{Ablation results of SCA on PAB.}
\vspace{-2mm}
\label{tab: ablation SCA}
\centering
\resizebox{0.33\textwidth}{!}{
\begin{tabular}{ccccc}
\bottomrule[1pt]
   \multicolumn{1}{c}{Methods} &R@1{$\uparrow$} &R@5{$\uparrow$} &R@10{$\uparrow$} \\
\bottomrule[0.5pt]
   \multicolumn{1}{c}{X-VLM}  &77.86 &98.28 &99.24 \\
   \multicolumn{1}{c}{X-VLM + SCA}  &\bf{80.54} &98.28  &99.24 \\
\bottomrule[1pt]
\end{tabular}}
\vspace{-2mm}
\end{table}

\subsection{Ablation Study}
\noindent\textbf{Baseline Model Selection}
In addition to X-VLM\cite{zeng2022multigrainedvisionlanguagepretraining}, we also tested two other models, EVA-CLIP~\cite{sun2023eva} and LLM2CLIP~\cite{huang2024llm2clippowerfullanguagemodel}, for Text-based Person Anomaly Search task. EVA-CLIP employs the 149M-parameter model from the EVA-02-CLIP series, while LLM2CLIP uses the fine-tuned Llama-3.1-1B as the text encoder. The image encoder for LLM2CLIP is based on the EVA CLIP series, specifically the EVA02-L-14-336 model with 428M parameters. 

Without fine-tuning, the $Recall@1$ of X-VLM has already reached 77.86, significantly surpassing EVA-CLIP's 60.01 and LLM-CLIP's 73.61. Following fine-tuning, X-VLM continues to outperform the other two methods. Therefore, we select X-VLM as our baseline model.

\noindent\textbf{Training Data} The training set of PAB~\cite{yang2024beyond} contains nearly 1 million normal and abnormal synthetic image-text pairs, which is an extremely large dataset. However, when fine-tuning X-VLM~\cite{zeng2022multigrainedvisionlanguagepretraining}, we did not use all the data but strategically selected a small part for fine-tuning. The reason for this choice is that we hope to fully optimize the model performance and reduce the time cost through the appropriate amount of data. As shown in Table~\ref{tab: data volume}, we can see that when only 10,000 pairs of image-text data pairs are used for fine-tuning, the effect is best, and the Recall@1 index even reaches 85.49. This result not only proves the importance of the right amount of data for model tuning but also shows that this choice is more effective than the strategy of fine-tuning with the full amount of data. With this approach, we can achieve higher accuracy in the image and text matching task while optimizing the efficiency of computing resources.

\noindent\textbf{Effectiveness of SCA} We conducted experiments on the unfine-tuned X-VLM model to verify the effectiveness of Similarity Cover Analysis (SCA). As shown in Table~\ref{tab: ablation SCA}, after introducing SCA, the $Recall@1$ of X-VLM is significantly improved by 2.68. This result shows that SCA plays an important role in optimizing model performance and can effectively improve the retrieval ability of the model.

\subsection{Visualization Results}
\begin{figure}[t]
  \centering
  \includegraphics[width=0.95\columnwidth]{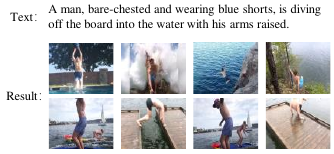}
  \caption{Visualisation of retrieval results on PAB.}
  \label{fig: Result}
\end{figure}

In Figure ~\ref{fig: Result}, we show the retrieval results of our method. In this example, our method is still able to retrieve the correct result when given a short text description. This performance stems from our model's enhanced capabilities in feature extraction and semantic understanding, ensuring efficient retrieval of matching answers regardless of the input information volume.

\section{Conclusion} \label{sec:conclusion}
In this paper, we propose a solution to the WWW 2025 challenge of Text-based Person Anomaly Search. Our method builds on the X-VLM model as a baseline and enhances the performance by fine-tuning it with a small amount of data. We also adopt a Similarity Cover Analysis strategy to improve retrieval capabilities. This approach significantly boosts the efficiency of identifying abnormal behaviors. Through comparative experiments with the PAB dataset, we achieve an accuracy of 85.49. This result not only demonstrates the effectiveness of our method but also provides new insights and technical means for the research on abnormal human behavior identification.

{\small
\bibliographystyle{ACM-Reference-Format}
\bibliography{egbib}
}

\end{CJK}
\end{document}